\pdfoutput=1
\documentclass{article}
\usepackage{xcolor}
\usepackage[preprint]{corl_2026}
\usepackage[T1]{fontenc}
\usepackage{amsmath,amssymb}
\usepackage{graphicx}
\usepackage{booktabs}
\usepackage{microtype}
\usepackage{enumitem}
\usepackage{tabularx}
\setlist{nosep, leftmargin=1.2em}
\graphicspath{{figures/out/}}
\hypersetup{pdftitle={Off-Policy RL with Online Inference Tuning: 1st Place Solution to the CoRL 2026 RoboEval Challenge},
            pdfauthor={Ilia Larchenko},
            pdfsubject={RoboEval Challenge, CoRL 2026},
            pdfkeywords={bimanual manipulation, VLA, off-policy reinforcement learning, behavioral metrics, Thompson sampling}}

\title{Off-Policy RL with Online Inference Tuning:\\ 1st Place Solution to the\\ CoRL 2026 RoboEval Challenge}

\author{
  Ilia Larchenko\\
  Independent Researcher\\
  \texttt{ilya.larchenko@gmail.com}
}

\begin{document}
\maketitle

\begin{abstract}
The CoRL 2026 RoboEval challenge has eight bimanual tasks, and it ranks a policy not only on whether it succeeds but on how: every stage of the task, done safely, efficiently and with both arms in coordination, scored against human demonstrations. My solution has two parts. The generic part, carried over from my LeHome challenge solution, is off-policy RL on a $\pi_{0.5}$-based policy that is also its own value function. It is trained with AWR and RECAP on advantages of the exact official score, in a loop where every rollout is at the same time a Thompson-sampling trial of about 20 inference parameters. Out of the box, it took the policy to 0.85 RDS on held-out seeds in two days (and it was still very far from convergence). The competition-specific part adds data that local exploration cannot find: generated demonstrations that reach the full reward, and recovery episodes that start from the policy's own failures. The submitted policy reaches 0.990 RDS on my held-out seeds and 0.987 on the official seeds, and it won first place in the challenge, with 0.988 in the organisers' final evaluation on 1{,}000 unpublished seeds per task.
\end{abstract}


\section{Introduction}

RoboEval~\citep{wang2025roboeval} asks for policies that are ``not just successful but well-behaved''. The challenge has eight bimanual tasks with two Franka arms (lifting a pot or a tray, stacking blocks, shelving or picking a book, a cube handover, packing a box, and rotating valves), each evaluated on 50 seeds with randomised positions and orientations; I train a single policy for all eight. The RoboEval Diagnostic Score (RDS) combines, per task, the outcome $O = 0.7\,\text{success} + 0.3\,\text{stage progress}$ with behavioral metrics (safety, efficiency and coordination) scored against the median human demonstration: $\text{RDS} = 0.4\,O + O\,(0.3\,S + 0.2\,E + 0.1\,C)$. A perfect score needs every stage of every task, every time, done at least as cleanly as a human does it.

I treated it as two problems. The \emph{generic} one is ``make a pretrained VLA better at an arbitrary score'', which my LeHome solution~\citep{larchenko2026lehome} already does: it optimises any target that can be computed, and its exploration acts exactly where this score is decided --- speed, smoothness, and how action chunks connect. The \emph{competition-specific} one is everything this recipe cannot reach: behavior that no demonstration shows and that is too far from the policy to be found by local exploration. Together they won first place in the challenge (Section~\ref{sec:final}).

\paragraph{What is in this report.} Section~\ref{sec:challenge} describes the challenge, Section~\ref{sec:method} the training loop, Section~\ref{sec:model} the model architecture and its inference, Section~\ref{sec:data} the data. Section~\ref{sec:story} follows the competition from start to finish: where the generic loop got me, what stopped it, and how I got from there to the final score. Section~\ref{sec:diag} has the results, Section~\ref{sec:final} the final results of the competition, and Appendix~\ref{app:v2} lists what is wrong with the demonstrations, with frames from them, and my suggestions for a RoboEval v2.

\section{The challenge}
\label{sec:challenge}

\begin{figure}[t]
  \centering
  \includegraphics[width=0.985\linewidth]{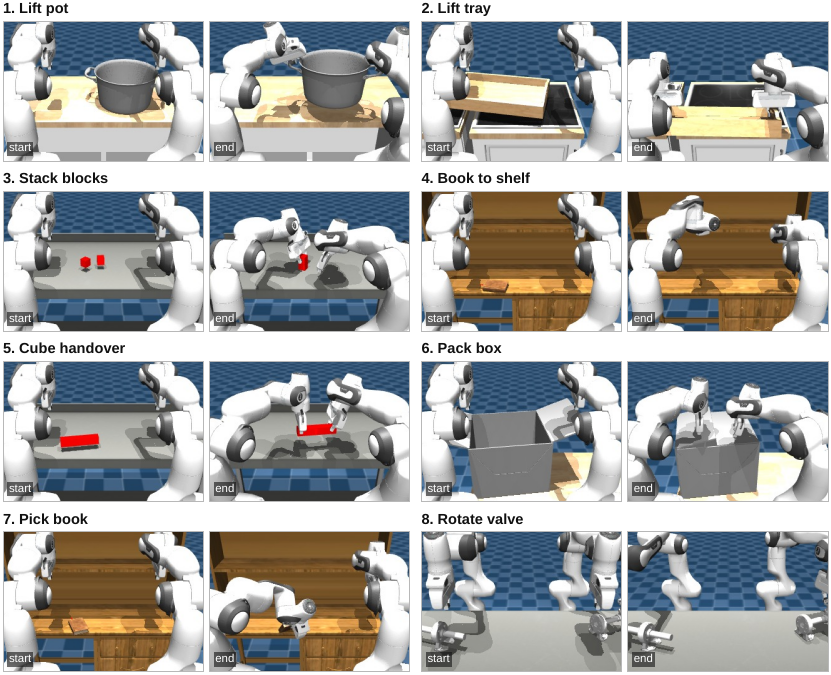}
  \caption{The eight tasks, each at the start and at the end of one full-reward episode (head camera; front
  camera for the valves, which the head camera does not see).}
  \label{fig:tasks}
\end{figure}

\begin{figure}[t]
  \centering
  \includegraphics[width=0.985\linewidth]{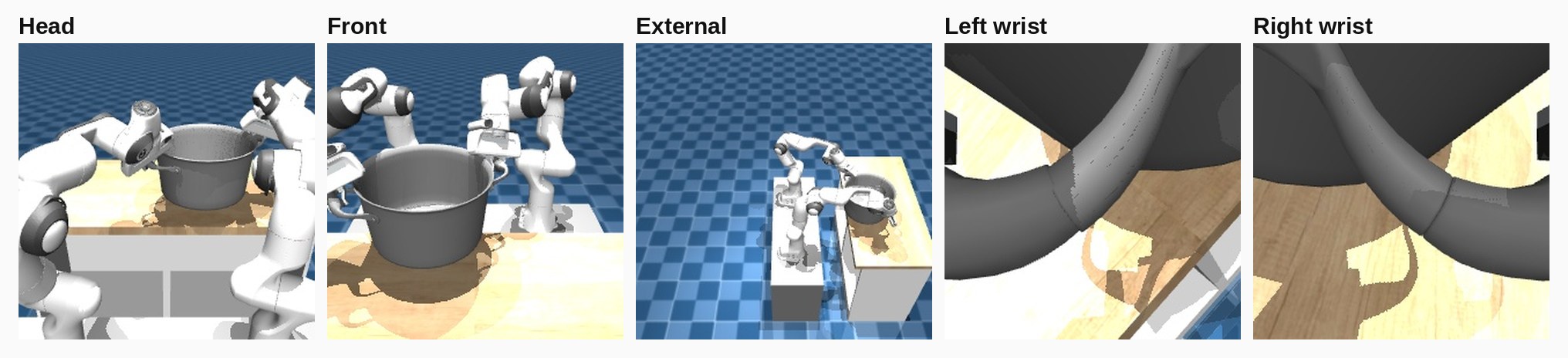}
  \caption{The five cameras the policy can use, here at the start of a LiftPot episode.}
  \label{fig:cameras}
\end{figure}

\paragraph{Setup.} The RoboEval CoRL 2026 challenge (Track 2 of the CoRL 2026 workshop challenges, organised by the RoboEval team at the University of Washington and AI2) runs in MuJoCo with a bimanual robot: two 7-DoF Franka Pandas with parallel grippers, mounted side by side (Fig.~\ref{fig:tasks}). There are eight task families (Table~\ref{tab:tasks}), all mandatory, and each is evaluated on its hardest variation, \emph{PositionAndOrientation}: the objects' positions and orientations are randomised by the seed. The evaluation uses 50 fixed, published seeds per task (1000--1049), the same for every team, so a submission is 400 episodes. An episode ends at success, at a failure condition, or at the task's time budget (10--40~s of simulated time). The policy acts at a fixed 20~Hz and may choose its action mode (joint positions or end-effector poses, absolute or delta; I use end-effector deltas) and its observations among the proprioception and five RGB cameras (head, front, external, and one on each wrist; Fig.~\ref{fig:cameras}); privileged simulator state (object poses) is forbidden at evaluation time, but synthetic data generation is allowed, as are any pretraining and any external data. The organisers provide 3{,}060 teleoperated demonstrations over 30 task variations, the RoboEval demonstration dataset~\citep{wang2025roboeval,roboeval2025dataset} (\href{https://huggingface.co/datasets/helen9975/roboeval_ee_delta_20hz_scaled}{\texttt{helen9975/roboeval\_ee\_delta\_20hz\_scaled}} on the Hugging Face Hub); for the eight evaluated ones that is 73--154 demonstrations per task.

\begin{table}[t]
  \centering\scriptsize
  \setlength{\tabcolsep}{3.2pt}
  \caption{The eight tasks as evaluated (PositionAndOrientation variation). Stages are the environment's own
  stage flags (stage progress = fraction reached); budget in simulated seconds. The last three columns are about the
  organisers' demonstrations of this variation only~\citep{roboeval2025dataset}, replayed under the challenge rules:
  demos = successful / all (the successful ones are the set the human reference medians come from); full = how many of
  them also reach every stage; demo RDS = the official score of all of them, as if they were a policy (mean 0.875).
  PackBox and StackTwoBlocks show the gap: almost every demonstration succeeds, almost none reaches every stage.}
  \label{tab:tasks}
  \begin{tabularx}{\linewidth}{@{}l>{\raggedright\arraybackslash}X>{\raggedright\arraybackslash}Xcccc@{}}
    \toprule
    Task & Success & Stages & Budget & Demos & Full & Demo RDS \\
    \midrule
    LiftPot & pot lifted 10~cm, level & left grasp $\cdot$ right grasp $\cdot$ lifted $\cdot$ lifted level, both holding & 10 & 72/77 & 68 & 0.914 \\
    LiftTray & heavy tray held by both hands, off the table & left grasp $\cdot$ right grasp $\cdot$ held clear of the table $\cdot$ lifted $\ge$2.3~cm & 10 & 100/154 & 100 & 0.698 \\
    StackTwoBlocks & one cube stacked on the other & a cube grasped $\cdot$ the other one grasped too $\cdot$ stacked, off the table & 20 & 135/141 & 1 & 0.830 \\
    StackSingleBookShelf & book on the shelf, released & grasp (after a push) $\cdot$ lift $\cdot$ touch the shelf $\cdot$ place and release & 40 & 66/73 & 66 & 0.909 \\
    CubeHandover & a rod passed from one hand to the other in the air & grasp $\cdot$ transfer $\cdot$ transfer with the rod off the table & 20 & 98/101 & 98 & 0.938 \\
    PackBox & both box flaps closed & left flap held $\cdot$ right flap held $\cdot$ right closed $\cdot$ left closed $\cdot$ box closed & 20 & 98/99 & 2 & 0.850 \\
    PickSingleBookFromTable & flat book picked up off the counter & grasp $\cdot$ lifted clear of the counter & 15 & 83/92 & 83 & 0.882 \\
    RotateValve & two handwheels turned $\ge$27$^\circ$ & valve 1 grasped $\cdot$ turned $\cdot$ valve 2 grasped $\cdot$ turned & 20 & 115/115 & 115 & 0.981 \\
    \bottomrule
  \end{tabularx}
\end{table}

\paragraph{The score in practice.} The evaluator records 14 metrics per episode: success and stage progress (outcome); self- and environment-collision counts, slips, joint and Cartesian jerk (safety); Cartesian, joint and orientation path length and completion time (efficiency; trajectory length is recorded but not scored, as it duplicates time at a fixed 20~Hz); the difference of the two arms' speeds and of the two grippers' heights (coordination). For each task it averages every metric over the 50 episodes and compares the mean to the median of the task's successful human demonstrations: at or better than the reference a metric scores 1, worse it scores $(r+c)/(v+c)$ with $c$ = 10\% of the reference (at least one event for the counts), so twice the reference scores about 0.5. A family score is the mean of its metrics, and the task score is the formula above; RDS is the mean over the eight tasks. Three consequences shaped my work:
\begin{itemize}
  \item \textbf{Behavior saturates at the human median.} A policy faster and smoother than the median demonstration gets exactly 1, and nothing more. Once a task's behavior is at the reference, only failures and stages are left to win.
  \item \textbf{Failures hurt more than anything else.} A failed episode loses its own outcome, and because the behavior terms are multiplied by the task's mean outcome, it also scales down the behavior score that all the other episodes earned. And since every metric is an average over the episodes, a failure that wanders around until the time budget runs out also inflates the task's mean path length and time.
  \item \textbf{Every stage counts}, including the ones the demonstrations skip (Section~\ref{sec:data}): 30\% of the outcome is stage progress, so completing the task in a non-canonical way (touching only one block, or not using the fingers to close the box) costs you some points.
\end{itemize}

\paragraph{Submission.} A submission is a self-contained bundle (under 5~GB): \texttt{metadata.yaml} (team, method, the declared interface: action mode, cameras, control frequency, policy seed), a short report, the checkpoint, and a \texttt{policy/} folder with the inference code (a \texttt{make\_policy} factory returning an object with \texttt{reset} and \texttt{act}) and its pinned requirements. The team runs the official evaluator itself, on its own hardware, on the official seeds; the evaluator writes \texttt{results.json} and per-episode logs, including the exact actions the policy sent. The portal re-validates every upload: it checks the schema, recomputes the aggregates from the logs and replays a sample of logged episodes in the simulator, which needs no checkpoint and must match the logged metrics. Each team may upload two submissions per day, and the most recent valid one is ranked. The rules forbid tuning on the published seeds, and the organisers re-run the top entries on unpublished seeds before awards (Section~\ref{sec:final}).

\paragraph{What this meant for my approach.} Three things followed. (i) The official seeds are only a final check: every decision was made on fresh random seeds of the same distribution (Section~\ref{sec:story}), which are free to generate, so I could afford hundreds of episodes per task per comparison. (ii) The score is computable for every episode I run, so it can be the reward itself (Section~\ref{sec:method}). (iii) Declared test-time compute (sampling, reranking) is allowed and the evaluator waits for the policy at every step, so best-of-$N$ sampling and guidance are cheap ways to buy score.

\section{Method: one loop that trains, explores and tunes}
\label{sec:method}

\begin{figure}[t]
  \centering
  \includegraphics[width=\linewidth]{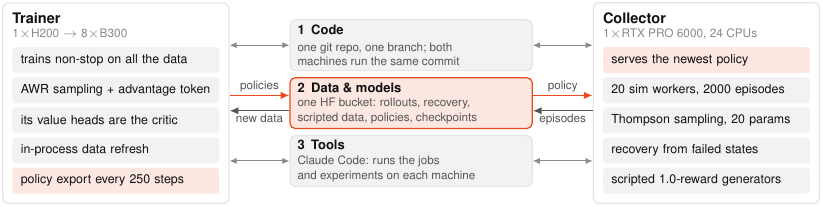}
  \caption{Two machines, integrated on three levels: they run the same commit, share data and models through one
  Hugging Face bucket, and I run the jobs on both with Claude Code.}
  \label{fig:system}
\end{figure}

\paragraph{The loop.} The training loop runs on two machines in parallel (Fig.~\ref{fig:system}). The trainer exports a policy every 250 steps. The collector serves the newest export to 20 MuJoCo workers and collects 2000 episodes per cycle (125 seeds $\times$ 2 per task); it also re-runs failed episodes from intermediate states (recovery) and, between collects, adds a batch of generated demonstrations (Section~\ref{sec:data}). The trainer picks every new batch up in-process, recomputes the advantages and swaps the data loader without stopping. Each dataset is sampled in proportion to its size times a share that decays with its age, so fresh on-policy data dominates. The two machines share \emph{code} (one repository and one branch), \emph{data and models} (one bucket) and \emph{operations}: a Claude Code~\citep{anthropic2025claudecode} agent on each machine runs its jobs. Data collection scales to several machines out of the box: I tested it with two collectors, but most of the time I used only one.

\begin{figure}[t]
  \centering
  \includegraphics[width=\linewidth]{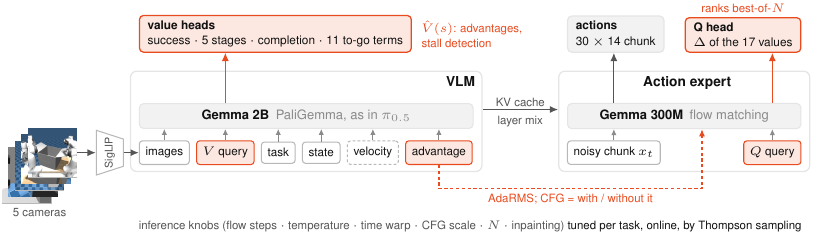}
  \caption{The policy: $\pi_{0.5}$~\citep{black2025pi05} plus a value query, a Q head and advantage
  conditioning (token + AdaRMS) that enables CFG.}
  \label{fig:model}
\end{figure}

\paragraph{The model is its own value function.} The architecture is inherited from my LeHome solution~\citep{larchenko2026lehome} and from our winning BEHAVIOR 2025 solution~\citep{larchenko2025behavior}: $\pi_{0.5}$ (a PaliGemma VLM and a flow-matching action expert, initialised from the public weights) with correlated flow noise, soft inpainting between chunks, and advantage conditioning through a token and AdaRMS (Fig.~\ref{fig:model}). A value query token reads the cameras and the task and predicts success, the stage indicators, completion, and the behavior terms of the score still to come. These predictions are recorded at every policy call during collection, so the critic used for the advantages is the policy that collected the data. A Q head on the action expert predicts how much a candidate action chunk changes those values, and picks the best of $N$ sampled chunks. Section~\ref{sec:model} has the details.

\paragraph{Advantages for an exact score.} The reward is the official RDS itself, linearised around the current policy: every step is credited with its change of each score metric $m$ (stage progress, path length, jerk, collisions, \ldots) times that metric's weight in the score, plus a small completion shaping. As in LeHome, the baseline is the model's own value, \emph{dampened} by how much it can be trusted:
\begin{align*}
  r_t &= \nabla_{\bar m}\,\text{RDS}\cdot\Delta m_t + \Phi_{t+1}-\Phi_t, &
  \delta_t &= r_t + \alpha\,(\hat V_{t+1}-\hat V_t),\\
  \alpha &= \mathrm{Cov}(G,\hat V)\,/\,\mathrm{Var}(\hat V), &
  A_t &= \textstyle\sum_{k\ge 0}\lambda^k\,\delta_{t+k},
\end{align*}
with completion potential $\Phi_t = 0.3\,\hat C_t$, $\alpha$ fitted per task on each new collect, GAE~\citep{schulman2016gae} with $\lambda = 0.95$, and $A$ normalised per task. The advantage is used twice: for \textbf{AWR}~\citep{peng2019awr} through the sampler, where frame $i$ is drawn with probability $\propto e^{\mathrm{clip}(A_i,-2,2)}$, and for \textbf{RECAP}~\citep{pi2025recap}, where the advantage token is on for positive frames and, at inference, classifier-free guidance (CFG)~\citep{ho2022cfg} pushes the action towards ``positive''.

\paragraph{An example of the linearised reward.} The gradient $\nabla_{\bar m}\text{RDS}$ is taken at the current policy's per-task mean metrics, so it says what an improvement is worth \emph{now}. On a task where the policy's mean path length is already below the human median, a shorter path is worth nothing (the metric is saturated at 1) and the reward ignores it; on StackTwoBlocks, where the path is longer than the reference, every centimetre saved is rewarded. Because the score is a ratio of means, the same collision is worth more on a task whose reference is 0 than on RotateValve, whose human median is 6 contacts with the valve.

\paragraph{Exploration and tuning in one go.} Every collected episode runs with its own draw of about 20 inference parameters, chosen per task by Thompson sampling~\citep{russo2018ts} on its official per-episode score: the number of flow steps, the noise temperature, correlated noise, the CFG scale, best-of-$N$, how many actions to execute and how many to carry into the next chunk, the execution speed, the gripper threshold, and switches that hide inputs. The same episodes are RL data, so the policy explores along exactly these axes, and the tuner keeps up with the policy as it changes. It also settles design questions empirically (Table~\ref{tab:decisions}). Speed is a good example, and one I also saw in LeHome: a policy trained on demonstrations moves more slowly than the demonstrations themselves, so early in RL the sampler prefers to play chunks 1.5--3.9$\times$ faster. RL then teaches the policy the right pace, and the preferred speed-up settles at about 1 on every task.

\paragraph{The tuner's model.} Each parameter is mapped to $[0,1]$ over its range, and the per-episode score (standardised per task) is modelled as additive smooth curves, $r = b_\text{task} + \sum_j f_j(u_j) + \sum_j g_{\text{task},j}(u_j) + \varepsilon$, each curve a weighted sum of Gaussian bumps: shared curves $f_j$ plus tighter per-task deviations $g_{\text{task},j}$, so tasks share evidence but may differ. This is Bayesian linear regression with a closed-form posterior updated per episode; a Thompson step draws all weights once and takes each parameter's argmax over its grid. Binary switches are Beta arms. When the policy changes, the posterior is refitted on all past episodes with recency weights.

\begin{table}[h]
  \centering\footnotesize
  \caption{Questions answered by the sampler (a difference below $\pm$0.02 per-episode RDS is noise).}
  \label{tab:decisions}
  \begin{tabular}{@{}lll@{}}
    \toprule
    Question & How it was tested & Answer \\
    \midrule
    Which inputs are needed? & input dropout + hide switches & state yes; velocities and front cameras no \\
    How much sampling noise? & temperature 0.1--1.2 & 0.1--0.55 \\
    How much guidance? & CFG scale 1--10 & 2.5--9.5 \\
    Best of how many chunks? & $N$ up to 6, ranked by the Q head & 2--6 \\
    Execute faster than trained? & speed 0.75--5$\times$ & first 1.5--3.9$\times$, now 0.8--1.1$\times$ \\
    \bottomrule
  \end{tabular}
\end{table}

\section{The model architecture}
\label{sec:model}

\paragraph{Backbone.} The VLM is PaliGemma~\citep{beyer2024paligemma} as in $\pi_{0.5}$~\citep{black2025pi05}: Gemma 2B with a SigLIP So400m vision tower, which I run with a 16-pixel patch on the native 256$\times$256 renders (the public /14 stem is resized at load), so each camera gives the same 16$\times$16 = 256 tokens; the whole vision tower is trained, and each camera has its own identity-initialised linear projection. The action expert is a 300M Gemma that attends to the VLM's key--value cache through a learned per-layer mix of the VLM layers (identity at initialisation) and is conditioned through AdaRMS on the flow time, the task and the advantage. There is no text prompt: the 257k-row Gemma vocabulary is never used.

\paragraph{Tokens.} The prefix (VLM) holds, in order:
\begin{itemize}
  \item \textbf{images}: 5 cameras $\times$ 256 tokens (a camera can be hidden: camera dropout in training, a per-task switch at inference);
  \item \textbf{$V$ query}: one learned token whose output feeds the value heads;
  \item \textbf{task}: one learned embedding per task (it plays the role of the language instruction);
  \item \textbf{state}: 2 continuous tokens (a linear map of the joint positions and grippers, and one of the velocities) and 30 per-dimension tokens (a 256-bin embedding of the value, plus a per-dimension embedding, plus a linear map of sin--cos features of the raw value);
  \item \textbf{advantage}: one learned token, present for positive-advantage frames;
  \item training only: the chunk's \textbf{FAST} tokens~\citep{pertsch2025fast} (an auxiliary next-token loss) and a \textbf{FAST-$Q$ query} that predicts the outcome of that tokenised chunk.
\end{itemize}
The suffix (action expert) holds the 30 noisy \textbf{action} tokens (14-D: per arm a 3-D position and a 3-D rotation vector relative to the chunk start, plus the two grippers) and one \textbf{$Q$ query}.

\begin{figure}[t]
  \centering
  \includegraphics[width=0.985\linewidth]{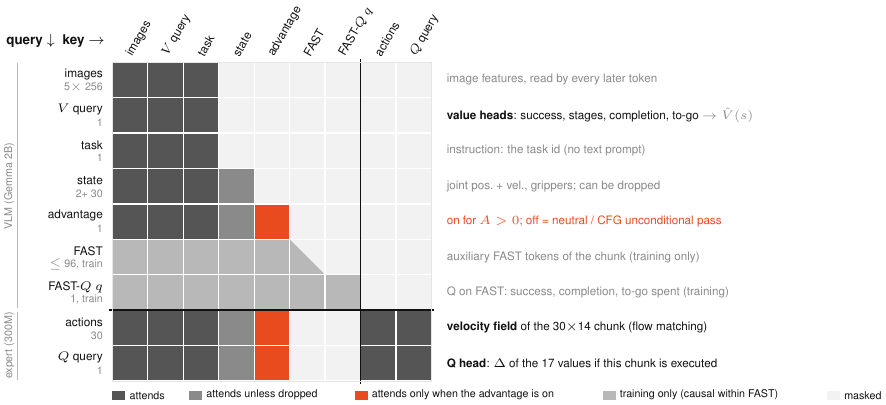}
  \caption{The attention mask by token group (rows attend to columns; groups not to scale). Right: what each
  query's output is used for.}
  \label{fig:attn}
\end{figure}

\paragraph{Attention mask.} Fig.~\ref{fig:attn} shows which groups attend to which. Three choices matter. (i) The $V$ query is in group 0, so the value is a function of the cameras and the task only: it never sees the advantage token (whose label it is later used to compute) or the state. (ii) The state and the advantage come after the images in their own groups: the image tokens are not modulated by them, and state, velocity or advantage tokens can be dropped per sample by masking only their columns. (iii) The $Q$ query sits in the action expert's group, bidirectional with the noisy actions, so it reads the candidate chunk at the current flow step together with the whole prefix. Classifier-free guidance needs no second prefix pass: the unconditional pass of the action expert masks the advantage column and drops the advantage AdaRMS term.

\paragraph{Heads.} All heads are linear layers on a query token's final hidden state:
\begin{itemize}
  \item on the $V$ query: \textbf{success} (BCE), \textbf{completion} ($t/T$ on successful episodes, 0 on failures, so it is progress times $P(\text{success})$), \textbf{stages} (for each of up to 5 stages, the probability that the episode reaches it; masked BCE) and \textbf{to-go} (the 11 behavior metrics still to accumulate until the end, normalised per task). The elapsed episode time enters only the heads (late fusion through a zero-initialised residual MLP; not the completion heads, which would otherwise read the clock), never the backbone or the actions;
  \item on the $Q$ query: how the candidate chunk changes the 17-D value vector $\psi$ = [success, 5 stages, 11 to-go], learned as a residual $\psi - \mathrm{sg}(\hat\psi(s_t))$ against the prefix's own stop-gradient estimate (i.e.\ $Q - V$), plus the completion 30 steps later; trained on the flow samples with $t \le 0.5$ (weighted by $1-t$) and read at the last denoising step;
  \item on the FAST-$Q$ query (training only): the same quantities predicted from the chunk's FAST tokens, i.e.\ a second, tokenised $Q$ estimate; it is an auxiliary loss whose gradients reach the VLM.
\end{itemize}
The losses are small next to the flow loss (weights 0.05--0.1 each), so the value heads shape the representation without dominating it.

\paragraph{Training details.} Advantage conditioning is inspired by RECAP~\citep{pi2025recap}, with many small differences in the details: a frame with $A<0$ is always trained neutral (token hidden, AdaRMS term off), a frame with $A \ge 0$ is neutral with a probability that falls from 0.5 at $A=0$ to 0.1 at $A \ge 2$, so the model learns both the conditional and the unconditional policy. Flow noise is correlated over the chunk, $\mathcal N(0, D(\beta C + (1-\beta)I)D)$ with $C$ the action-chunk correlation and $D$ the per-timestep scale; $\beta$ is chosen per task at inference. State, velocity and camera dropout during training are what make the hide switches of Table~\ref{tab:decisions} possible at inference.

\begin{figure}[t]
  \centering
  \includegraphics[width=0.985\linewidth]{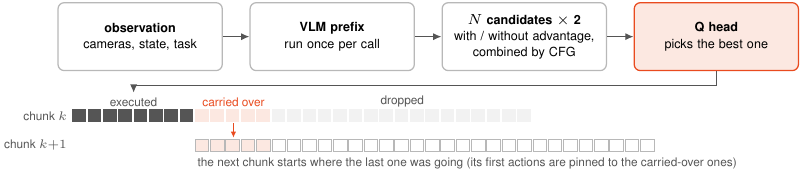}
  \caption{One policy call (top) and how consecutive chunks connect (bottom). Each of the $N$ candidates is denoised with and without the advantage input, and CFG combines the two. The Q head picks one
  candidate; its first actions are executed, the next few are carried over into the next chunk, the rest is
  dropped. All test-time parameters are optimised per task (Table~\ref{tab:params}).}
  \label{fig:infer}
\end{figure}

\paragraph{Inference.} Fig.~\ref{fig:infer} shows one call. The prefix runs once; the $N$ candidates (6 in the bundle; each task uses the best of its first $k \le 6$) start from independent correlated noise scaled by a temperature and run 5--19 flow steps on a warped time grid, all candidates and the CFG pass in one batch. Each candidate is scored by the Q head: by its predicted change of success, or, on three tasks, by the score-weighted change of all 17 values, and the best is executed. Inpainting~\citep{black2025rtc} connects chunks: the first \emph{keep} actions of the new chunk are pinned (softly, or hard with \emph{snap}) to the part of the previous chunk that was planned but not executed, for the first flow steps only, so the new chunk starts where the old one was going. The executed part is resampled to \emph{speed} (a cubic spline over the poses: 0.86 plays 7 planned actions over 8 steps), and the gripper command is binarised at a per-task threshold. Finally, a \emph{stall boost} (all tasks except CubeHandover): when the success prediction drops or stays flat, the next three calls use a 1.5$\times$ hotter noise and the best of all $N$ candidates. Table~\ref{tab:params} lists the final per-task values. They differ a lot between tasks (5 flow steps on the book tasks, 19 on LiftPot; executing 7 actions per call on CubeHandover, 25 on PackBox), and none of them was set by hand.

\begin{table}[t]
  \centering\scriptsize
  \setlength{\tabcolsep}{3.3pt}
  \caption{Final per-task inference parameters. $k$: best of the first $k$ of $N=6$
  candidates; rank: by $\Delta$success or by the score-weighted $\Delta$ of all values.}
  \label{tab:params}
  \begin{tabular}{@{}lcccccccccl@{}}
    \toprule
    Task & $k$ & rank & flow steps & temp. & CFG & replan & keep & speed & gripper & hidden \\
    \midrule
    LiftPot & 5 & succ. & 19 & 0.10 & 9.4 & 8 & 5 & 1.00 & 0.90 & front, external \\
    LiftTray & 3 & succ. & 11 & 0.10 & 4.5 & 24 & 6 & 1.05 & 0.89 & front, external \\
    StackTwoBlocks & 6 & score & 15 & 0.21 & 4.7 & 9 & 0 & 1.05 & 0.60 & external, velocity \\
    StackSingleBookShelf & 6 & succ. & 5 & 0.15 & 4.7 & 7 & 6 & 0.86 & 0.57 & front, velocity \\
    CubeHandover & 5 & succ. & 17 & 0.35 & 4.7 & 7 & 0 & 0.91 & 0.69 & -- \\
    PackBox & 5 & score & 7 & 0.36 & 6.3 & 25 & 5 & 0.95 & 0.66 & external \\
    PickSingleBookFromTable & 2 & succ. & 5 & 0.18 & 5.0 & 18 & 5 & 0.91 & 0.65 & front, ext., velocity \\
    RotateValve & 4 & score & 5 & 0.40 & 2.5 & 17 & 0 & 1.00 & 0.60 & front, external \\
    \bottomrule
  \end{tabular}
\end{table}

\section{Data: what the demonstrations never show}
\label{sec:data}

\paragraph{Demonstrations.} The organisers' demonstrations~\citep{roboeval2025dataset} are where the policy starts, and on some tasks they also cap it. In StackTwoBlocks, one stage needs both blocks held in the air at once, which only one of the 141 demonstrations does; in PackBox, the flaps have to be held with the fingers, while the demonstrations close them with the hand (2 of 99 reach every stage); on LiftTray, a third of the recordings end before the tray is lifted far enough to count. Local exploration cannot discover strategies like these. Appendix~\ref{app:demos} goes through the problems task by task, with frames from the demonstrations.

\paragraph{Generated demonstrations.} So I generated new demonstrations. For each task, a scripted policy reads the object poses from the simulator and executes a smooth, state-parametrised strategy close to the demonstrated one but complete: both blocks held together, flaps pressed with closed fingers, books pushed to an overhang before the pinch. The scripts were built in an agentic auto-research loop with Claude Code: run a few hundred episodes on separate test seeds, read the per-stage and per-metric breakdown of the failures, change the strategy, and repeat. They reach the full reward in more than 99.5\% of episodes, with RDS 0.997--1.000 on all tasks except PackBox. This approach cannot be submitted as a policy, but it can generate synthetic training data: every collector cycle adds a generation of full-reward episodes (500--1000 per task), trained at advantage 1.0 and with extra state dropout, so that the policy learns from the cameras rather than copying the proprioception. They get a large weight because the policy, already trained in the demonstrations' style, has to be retrained (see the plateau in Fig.~\ref{fig:results}); training on them from the start would probably need much less. In total I generated around 200k synthetic trajectories, but some of them had issues and, as the logic changed over time, many of them were discarded.

\paragraph{Why scripts, and why state-parametrised.} A scripted controller with privileged state is the cheapest way to show the policy a strategy it cannot find, but a fixed script would teach one rigid trajectory. Each script therefore draws its own parameters per episode (speeds, offsets, grasp points, waits within ranges) and reacts to what happens (a regrasp when a close misses, a re-push when the book's overhang is too short), and every decision is a function of the object poses, so the policy sees a family of strategies that it can only reproduce from the cameras. The scripts also had to respect the behavior metrics the policy is scored on: for example, the CubeHandover script keeps the receiving hand at the giving hand's height (gripper height difference 0.01 against the human 0.037), and the PackBox script opens its fingers while lifting off a flap, because a hold lost while the gripper is not opening counts as a slip.

\paragraph{Rollouts and recovery.} On top of the scripted demonstrations, the policy produced more than 240k rollouts of its own over the whole competition, each with its recorded value predictions. All failed and partially failed episodes go through a hierarchical recovery: the episode is replayed to a branch point (the failure, 2~s before it, or the last value peak) and continued from there, first by the policy itself and then by the scripted policy, which either re-attempts the task from the failed state or steers the episode back onto its own path. Only continuations that reach the full reward are kept, about 100 per collect. The only external pretraining is the public $\pi_{0.5}$ base model.

\section{Interface and compute}

The policy controls both arms and grippers with end-effector delta actions at 20~Hz. It predicts 30-step action chunks (1.5~s) as poses relative to the commanded pose at the chunk start, which convert exactly to per-step deltas. It observes five RGB cameras at 256$\times$256 and the proprioception (joint positions and velocities, gripper states). Test-time compute (best-of-$N$ with CFG, Section~\ref{sec:model}) is batched, so one call stays cheap.

\paragraph{Compute.} The trainer used one H200 for about the first half of the time and eight B300s for the second half: about 600 GPU-hours. The collector is a single RTX PRO 6000 with 24 CPU cores, running continuously.

\section{Competition progression}
\label{sec:story}

\begin{figure}[t]
  \centering
  \includegraphics[width=0.985\linewidth]{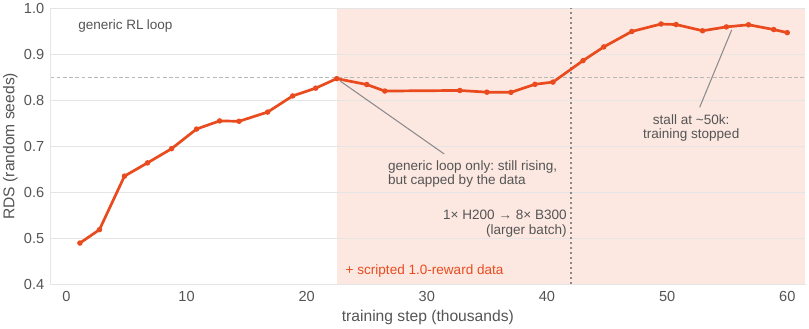}
  \caption{I worked on the competition for around 9 days; the curve is the RDS of every collect (2000
  episodes on held-out seeds, each with its own draw of inference parameters; mean per 2k steps) over the
  policy's training steps, up to step 60k, where I stopped training. At step 42k training moved from one H200 to
  eight B300s with a larger batch (dashed line).}
  \label{fig:results}
\end{figure}

\paragraph{The generic loop alone.} I started by taking my LeHome solution as it was, with no scripted data at all: only the organisers' demonstrations and the policy's own rollouts. In about two days of training on one GPU it got to 0.85 RDS (Fig.~\ref{fig:results}). That number comes from the collects, where every episode explores the inference parameters, so with the best parameters it was probably higher, and it was still going up. Some tasks were already above 0.95 and heading towards 1.

\paragraph{Where the data stops.} At that point I did the first failure analysis and found that on some tasks the demonstrations would not let the policy get anywhere near 1.0 (Section~\ref{sec:data} and Appendix~\ref{app:demos}). My loop explores a lot in speed, smoothness and how chunks connect, plus some local exploration around the actions, but it cannot invent a new strategy: if no demonstration holds both blocks in the air, no rollout close to the demonstrations will either, and RL can only credit a stage that some rollout reaches. On a couple of tasks a perfect score is very hard or even impossible anyway: the StackTwoBlocks path reference comes from demonstrations that do everything with one arm, and when you properly use both hands the total path is much longer than in those demonstrations; and in some PackBox seeds the robot touches the cabinet during the reset, before the policy acts (Appendix~\ref{app:v2}).

\paragraph{Scripted data.} So I wrote scripted policies that read the simulator state, generated clean full-reward demonstrations, and mixed a new batch of them in at every loop (Section~\ref{sec:data}). It took a while to pay off: the policy had learned the demonstrators' style and had to relearn the behavior, so the score first dipped and plateaued. After the plateau it started to improve again, and this time without a ceiling.

\paragraph{The stall.} Around step 50k progress stalled at about 0.97 in the collects (about 0.98 on held-out evaluations with the best parameters), and further training up to step 60k did not move it (the end of the curve in Fig.~\ref{fig:results}).

\paragraph{Final optimisation.} For the last two days I stopped training and focused on the final model. I took a few checkpoints from around step 50k and ran a very extensive search over inference parameters and weight averages~\citep{wortsman2022soups}. The submitted model averages the weights of two checkpoints ($0.65\,\theta_{48250} + 0.35\,\theta_{58000}$) and uses the best set of parameters for each task (Table~\ref{tab:params}). Evaluation is fast and cheap, so every option could be tested on new held-out random seeds, which kept me from overfitting to the seeds I had used to choose it. All of this effort gave about +0.01 RDS: the best single checkpoint with its own tuned parameters scored 0.981 on 500 held-out episodes per task, and the final model 0.990 on another 500 (Table~\ref{tab:heldout}). That matters on a leaderboard; in a real deployment it would probably not be worth the effort.

\paragraph{Evaluation discipline.} Every decision was a paired comparison on fresh random seeds of the official distribution; the official seeds (1000--1049) were used only to produce the submissions. Selection bias is real at this level: with dozens of candidates and differences of 0.005, the best option on the selection seeds is often just the luckiest one. So an option was adopted only after it won again on seeds that played no part in choosing it.

\paragraph{Noise and reproducibility.} Identical settings do not reproduce exactly: GPU numerics differ between batch compositions, and the evaluator runs several workers sharing one GPU, so the same bundle on the same seeds gives slightly different episodes. A single 50-seed run varies by about $\pm$0.005 RDS, so I treated an official run as one noisy sample, never as a measurement to tune against.

\section{Results}
\label{sec:diag}

\begin{table}[t]
  \centering\footnotesize
  \setlength{\tabcolsep}{4pt}
  \caption{The submitted policy on held-out seeds: 500 episodes per task (fresh random seeds, each task's final
  inference parameters). Success; full = success with every stage reached; then the outcome, the three behavior
  family scores and the task score (RDS).}
  \label{tab:heldout}
  \begin{tabular}{@{}lccccccc@{}}
    \toprule
    Task & Success & Full & Outcome & Safety & Efficiency & Coord. & Score \\
    \midrule
    LiftPot & 1.000 & 0.986 & 0.999 & 1.000 & 1.000 & 1.000 & 0.999 \\
    LiftTray & 0.998 & 0.998 & 0.998 & 1.000 & 1.000 & 1.000 & 0.998 \\
    StackTwoBlocks & 0.996 & 0.994 & 0.997 & 0.993 & 0.920 & 1.000 & 0.979 \\
    StackSingleBookShelf & 0.998 & 0.998 & 0.998 & 0.999 & 1.000 & 1.000 & 0.998 \\
    CubeHandover & 1.000 & 1.000 & 1.000 & 0.993 & 1.000 & 1.000 & 0.998 \\
    PackBox & 1.000 & 0.990 & 0.999 & 0.918 & 1.000 & 1.000 & 0.974 \\
    PickSingleBookFromTable & 0.988 & 0.988 & 0.990 & 0.995 & 0.995 & 1.000 & 0.988 \\
    RotateValve & 0.988 & 0.980 & 0.989 & 1.000 & 1.000 & 1.000 & 0.989 \\
    \midrule
    Mean & 0.996 & 0.992 & 0.996 & 0.987 & 0.989 & 1.000 & 0.990 \\
    \bottomrule
  \end{tabular}
\end{table}

\begin{table}[t]
  \centering\footnotesize
  \setlength{\tabcolsep}{4pt}
  \caption{The submitted policy on the official seeds: 50 episodes per task. Five of the eight tasks score exactly 1.000.}
  \label{tab:official}
  \begin{tabular}{@{}lccccccc@{}}
    \toprule
    Task & Success & Full & Outcome & Safety & Efficiency & Coord. & Score \\
    \midrule
    LiftPot & 1.000 & 1.000 & 1.000 & 1.000 & 1.000 & 1.000 & 1.000 \\
    LiftTray & 0.980 & 0.980 & 0.984 & 1.000 & 1.000 & 1.000 & 0.984 \\
    StackTwoBlocks & 1.000 & 1.000 & 1.000 & 0.978 & 0.919 & 1.000 & 0.977 \\
    StackSingleBookShelf & 1.000 & 1.000 & 1.000 & 1.000 & 1.000 & 1.000 & 1.000 \\
    CubeHandover & 1.000 & 1.000 & 1.000 & 1.000 & 1.000 & 1.000 & 1.000 \\
    PackBox & 1.000 & 1.000 & 1.000 & 0.898 & 1.000 & 1.000 & 0.969 \\
    PickSingleBookFromTable & 0.960 & 0.960 & 0.972 & 0.992 & 0.984 & 1.000 & 0.967 \\
    RotateValve & 1.000 & 1.000 & 1.000 & 1.000 & 1.000 & 1.000 & 1.000 \\
    \midrule
    Mean & 0.993 & 0.993 & 0.995 & 0.983 & 0.988 & 1.000 & 0.987 \\
    \bottomrule
  \end{tabular}
\end{table}

The submitted policy reaches 0.990 RDS on held-out seeds (Table~\ref{tab:heldout}) and \textbf{0.987} RDS with 99.25\% success on the official seeds (Table~\ref{tab:official}).

\paragraph{Official run.} Table~\ref{tab:official} breaks the official 0.987 down. Five tasks are perfect; the losses are two failures in PickSingleBookFromTable and one in LiftTray (outcome), the StackTwoBlocks path length (efficiency, see Appendix~\ref{app:v2}), and the PackBox environment collisions (safety). Held-out (0.990, 500 episodes per task) and official (0.987, 50) agree within the official run's noise. The organisers' final evaluation, on their own unpublished seeds, is in Section~\ref{sec:final}.

\paragraph{Diagnostic analysis.} Coordination is 1.00 on every task, and every task succeeds in more than 98\% of held-out episodes: real failures are rare. Most of the remaining loss is behavioral: StackTwoBlocks efficiency (its reference comes from one-arm demos) and PackBox safety (in some seeds the robot touches the cabinet at reset, before the policy acts).

\paragraph{Discussion.} The generic part of the method, off-policy RL on a policy that is its own value function, with advantages of the exact score and inference parameters tuned in the same loop, does not depend on this benchmark: it needs only rollouts and a score, and it took the policy from the demonstrations to at least 0.85 RDS in just a couple of days of training on one GPU, without any task-specific engineering. Squeezing out the rest of the score was not data-efficient, though: it took many more rollouts and scripted demonstrations to get past the limits of the data. It applies anywhere rollouts are affordable. The privileged scripted demonstrations are a different kind of tool: they work when the simulator (or a reliable real-to-sim-to-real pipeline) is trustworthy and the task is easy to script and cheap to run. That is not true of every task, but where these conditions hold it is a useful tool. They also had a cost I did not measure but could see: after the switch to the much cleaner scripted data, the policy visibly lost some of the recoveries (regrasps, retries) it had learned from its own messy rollouts.

\paragraph{Code, checkpoint and videos.} The code is at \url{https://github.com/IliaLarchenko/RoboEvalSolution} and the checkpoint, a fine-tune of $\pi_{0.5}$ (openpi), at \url{https://huggingface.co/IliaLarchenko/RoboEvalSolution}; both are released under Apache-2.0. The blog post with videos of the policy is at \url{https://ilialarchenko.com/projects/roboeval2026/}.

\section{Competition results}
\label{sec:final}

\begin{table}[!ht]
  \centering\footnotesize
  \setlength{\tabcolsep}{4pt}
  \caption{The final results of the challenge, as published by the organisers: the ranking is by RDS on 1{,}000 unpublished seeds per task. The last column is the leaderboard score on the 50 official seeds per task. The first row is the policy in this report.}
  \label{tab:final}
  \begin{tabularx}{\linewidth}{@{}cl>{\raggedright\arraybackslash}Xccc@{}}
    \toprule
    & & & \multicolumn{2}{c}{Unpublished seeds} & Official seeds \\
    \cmidrule(lr){4-5}\cmidrule(l){6-6}
    Rank & Team & Method & RDS & Success & RDS \\
    \midrule
    \textbf{1} & \textbf{Ilia Larchenko} & VLA with off-policy RL and inference tuning & \textbf{0.988} & \textbf{99.3\%} & 0.987 \\
    2 & Metric AI Lab & RDB-ChunkFlow & 0.982 & 99.1\% & 0.986 \\
    3 & LocalModels & ACT-bimanual & 0.978 & 98.9\% & 0.987 \\
    \bottomrule
  \end{tabularx}
\end{table}

The final results were announced on November 12, 2026, at the CoRL workshop, and my policy took \textbf{first place} (Table~\ref{tab:final}). On the 50 official seeds per task the three leading entries were within 0.002 of each other, well inside the noise of a 50-seed run. So the organisers re-evaluated all three, each exactly as submitted, on a new set of 1{,}000 unpublished seeds per task, the same for every team and fixed before any run, and ranked them by the RDS on these seeds. My policy scored 0.988 with 99.3\% success, 0.006 ahead of the second place and 0.010 ahead of the third; the organisers report that every pairwise difference is significant (the 95\% paired bootstrap interval over seeds excludes zero). The score also agrees with my own held-out evaluation (0.990 on 500 episodes per task, Table~\ref{tab:heldout}). The official results are at \url{https://robo-eval.github.io/challenge.html#results}.

\bibliography{references}

\begin{thebibliography}{15}
\providecommand{\natexlab}[1]{#1}
\providecommand{\url}[1]{\texttt{#1}}
\expandafter\ifx\csname urlstyle\endcsname\relax
  \providecommand{\doi}[1]{doi: #1}\else
  \providecommand{\doi}{doi: \begingroup \urlstyle{rm}\Url}\fi

\bibitem[Wang et~al.(2025)Wang, Ung, Tan, Tannert, Duan, Li, Le, Oswal, Grotz,
  Pumacay, Deng, Krishna, Fox, and Srinivasa]{wang2025roboeval}
Y.~R. Wang, C.~Ung, C.~Tan, G.~Tannert, J.~Duan, J.~Li, A.~Le, R.~Oswal,
  M.~Grotz, W.~Pumacay, Y.~Deng, R.~Krishna, D.~Fox, and S.~Srinivasa.
\newblock {RoboEval}: Where robotic manipulation meets structured and scalable
  evaluation.
\newblock \emph{arXiv preprint arXiv:2507.00435}, 2025.
\newblock URL \url{https://arxiv.org/abs/2507.00435}.

\bibitem[Larchenko(2026)]{larchenko2026lehome}
I.~Larchenko.
\newblock Learning to fold: Prizewinning solution at {LeHome} challenge 2026
  (1st place online, 2nd offline).
\newblock \emph{arXiv preprint arXiv:2606.27163}, 2026.
\newblock URL \url{https://arxiv.org/abs/2606.27163}.

\bibitem[{RoboEval team}(2026)]{roboeval2025dataset}
{RoboEval team}.
\newblock {RoboEval} demonstration dataset
  (\texttt{roboeval\_ee\_delta\_20hz\_scaled}).
\newblock Hugging Face Hub, 2026.
\newblock URL
  \url{https://huggingface.co/datasets/helen9975/roboeval_ee_delta_20hz_scaled}.
\newblock Released with the RoboEval CoRL 2026 challenge (accompanies
  \citet{wang2025roboeval}); accessed October 2026.

\bibitem[{Anthropic}(2025)]{anthropic2025claudecode}
{Anthropic}.
\newblock {Claude Code}, 2025.
\newblock URL \url{https://claude.com/product/claude-code}.

\bibitem[{Physical Intelligence} et~al.(2025){Physical Intelligence}, Black,
  et~al.]{black2025pi05}
{Physical Intelligence}, K.~Black, et~al.
\newblock $\pi_{0.5}$: a vision-language-action model with open-world
  generalization.
\newblock \emph{arXiv preprint arXiv:2504.16054}, 2025.
\newblock URL \url{https://arxiv.org/abs/2504.16054}.

\bibitem[Larchenko et~al.(2025)Larchenko, Zarin, and
  Karnatak]{larchenko2025behavior}
I.~Larchenko, G.~Zarin, and A.~Karnatak.
\newblock Task adaptation of vision-language-action model: 1st place solution
  for the 2025 {BEHAVIOR} challenge.
\newblock \emph{arXiv preprint arXiv:2512.06951}, 2025.
\newblock URL \url{https://arxiv.org/abs/2512.06951}.

\bibitem[Schulman et~al.(2016)Schulman, Moritz, Levine, Jordan, and
  Abbeel]{schulman2016gae}
J.~Schulman, P.~Moritz, S.~Levine, M.~Jordan, and P.~Abbeel.
\newblock High-dimensional continuous control using generalized advantage
  estimation.
\newblock In \emph{International Conference on Learning Representations
  (ICLR)}, 2016.
\newblock URL \url{https://arxiv.org/abs/1506.02438}.

\bibitem[Peng et~al.(2019)Peng, Kumar, Zhang, and Levine]{peng2019awr}
X.~B. Peng, A.~Kumar, G.~Zhang, and S.~Levine.
\newblock Advantage-weighted regression: Simple and scalable off-policy
  reinforcement learning.
\newblock \emph{arXiv preprint arXiv:1910.00177}, 2019.
\newblock URL \url{https://arxiv.org/abs/1910.00177}.

\bibitem[{Physical Intelligence} et~al.(2025){Physical Intelligence}, Amin,
  et~al.]{pi2025recap}
{Physical Intelligence}, A.~Amin, et~al.
\newblock $\pi^{*}_{0.6}$: a {VLA} that learns from experience.
\newblock \emph{arXiv preprint arXiv:2511.14759}, 2025.
\newblock URL \url{https://arxiv.org/abs/2511.14759}.

\bibitem[Ho and Salimans(2022)]{ho2022cfg}
J.~Ho and T.~Salimans.
\newblock Classifier-free diffusion guidance.
\newblock \emph{arXiv preprint arXiv:2207.12598}, 2022.
\newblock URL \url{https://arxiv.org/abs/2207.12598}.

\bibitem[Russo et~al.(2018)Russo, Van~Roy, Kazerouni, Osband, and
  Wen]{russo2018ts}
D.~J. Russo, B.~Van~Roy, A.~Kazerouni, I.~Osband, and Z.~Wen.
\newblock A tutorial on {Thompson} sampling.
\newblock \emph{Foundations and Trends in Machine Learning}, 11\penalty0
  (1):\penalty0 1--96, 2018.
\newblock URL \url{https://arxiv.org/abs/1707.02038}.

\bibitem[Beyer et~al.(2024)Beyer, Steiner, Pinto, Kolesnikov, Wang, Salz,
  Neumann, Alabdulmohsin, Tschannen, Bugliarello, et~al.]{beyer2024paligemma}
L.~Beyer, A.~Steiner, A.~S. Pinto, A.~Kolesnikov, X.~Wang, D.~Salz, M.~Neumann,
  I.~Alabdulmohsin, M.~Tschannen, E.~Bugliarello, et~al.
\newblock {PaliGemma}: A versatile {3B} {VLM} for transfer.
\newblock \emph{arXiv preprint arXiv:2407.07726}, 2024.
\newblock URL \url{https://arxiv.org/abs/2407.07726}.

\bibitem[Pertsch et~al.(2025)Pertsch, Stachowicz, Ichter, Driess, Nair, Vuong,
  Mees, Finn, and Levine]{pertsch2025fast}
K.~Pertsch, K.~Stachowicz, B.~Ichter, D.~Driess, S.~Nair, Q.~Vuong, O.~Mees,
  C.~Finn, and S.~Levine.
\newblock {FAST}: Efficient action tokenization for vision-language-action
  models.
\newblock \emph{arXiv preprint arXiv:2501.09747}, 2025.
\newblock URL \url{https://arxiv.org/abs/2501.09747}.

\bibitem[Black et~al.(2025)Black, Galliker, and Levine]{black2025rtc}
K.~Black, M.~Y. Galliker, and S.~Levine.
\newblock Real-time execution of action chunking flow policies.
\newblock In \emph{Advances in Neural Information Processing Systems
  (NeurIPS)}, 2025.
\newblock URL \url{https://arxiv.org/abs/2506.07339}.

\bibitem[Wortsman et~al.(2022)Wortsman, Ilharco, Gadre, Roelofs, Gontijo-Lopes,
  Morcos, Namkoong, Farhadi, Carmon, Kornblith, and Schmidt]{wortsman2022soups}
M.~Wortsman, G.~Ilharco, S.~Y. Gadre, R.~Roelofs, R.~Gontijo-Lopes, A.~S.
  Morcos, H.~Namkoong, A.~Farhadi, Y.~Carmon, S.~Kornblith, and L.~Schmidt.
\newblock Model soups: Averaging weights of multiple fine-tuned models improves
  accuracy without increasing inference time.
\newblock In \emph{International Conference on Machine Learning (ICML)}, volume
  162 of \emph{PMLR}, pages 23965--23998, 2022.
\newblock URL \url{https://proceedings.mlr.press/v162/wortsman22a.html}.

\end{thebibliography}

\appendix
\section{Toward RoboEval v2}
\label{app:v2}

I really enjoyed this challenge. RoboEval scores how a task is done and not only whether it is done, it is cheap to run, and its stage flags and behavior metrics tell you exactly where a policy loses points. Thanks a lot to the organisers for building it and running the competition. Several entries are now close to the maximum, so here is what I would change in a second version, starting with the data.

\subsection{What is wrong with the demonstrations}
\label{app:demos}
\begin{figure}[t]
  \centering
  \begin{minipage}{0.97\linewidth}\centering
  \includegraphics[width=0.97\linewidth]{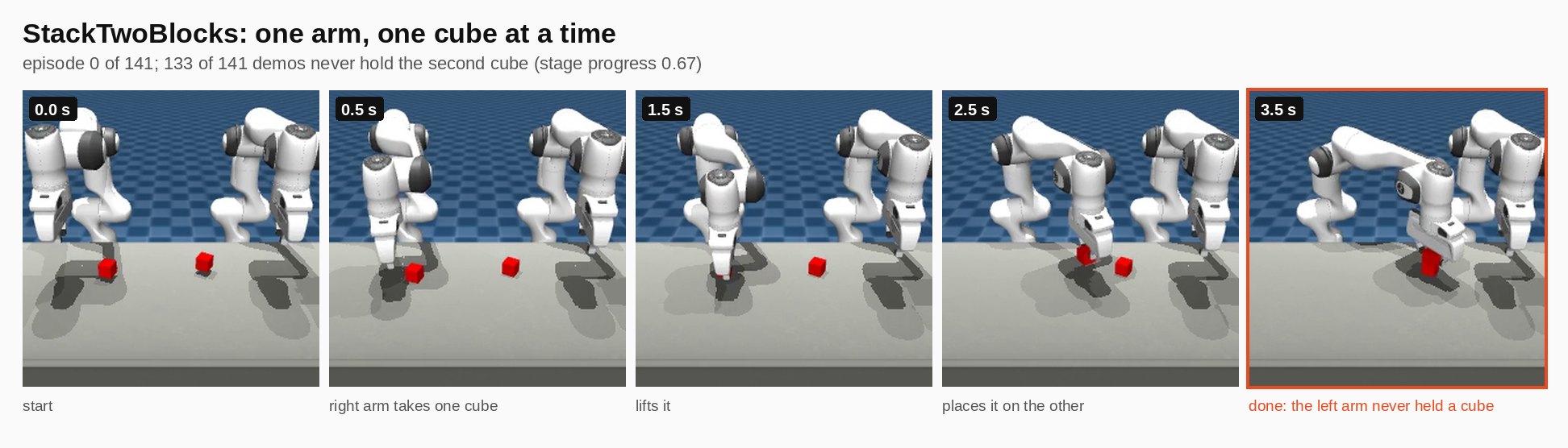}\\[4pt]
  \includegraphics[width=0.97\linewidth]{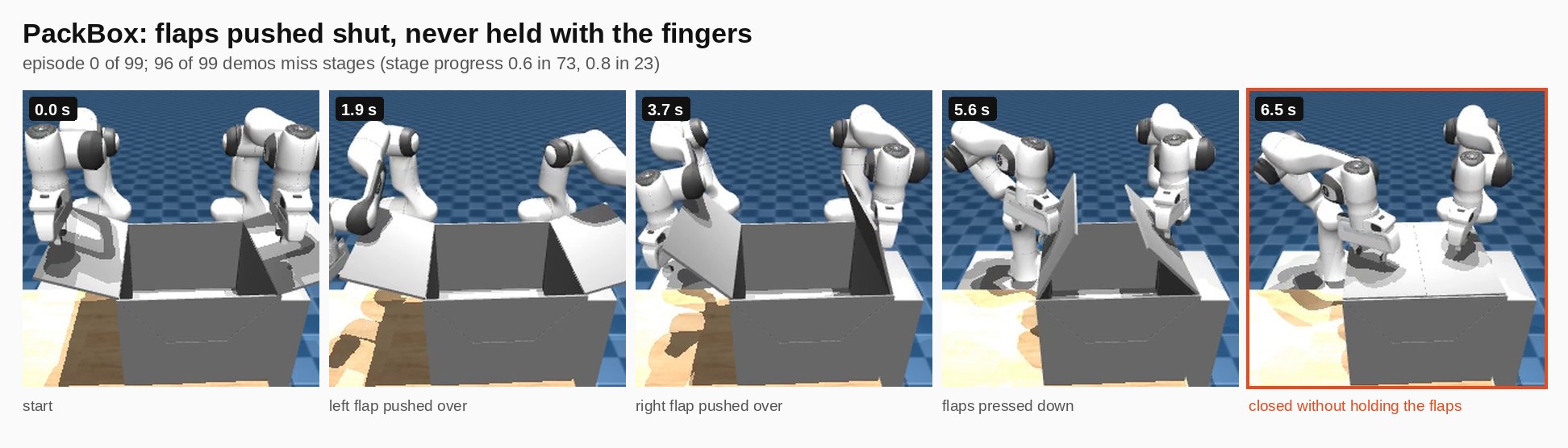}\\[4pt]
  \includegraphics[width=0.97\linewidth]{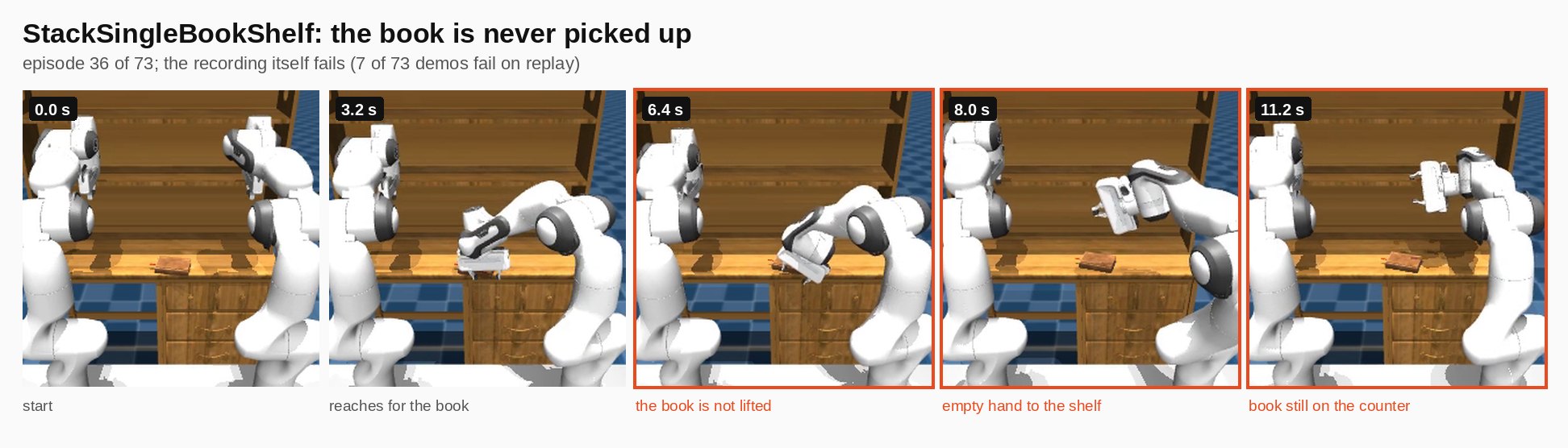}
  \end{minipage}
  \caption{What goes wrong in the demonstrations: StackTwoBlocks, PackBox and StackSingleBookShelf. Five frames from one demonstration each, from the dataset's own recordings; the frames that show the problem are outlined.}
  \label{fig:demos}
\end{figure}

I replayed every demonstration of the evaluated variation under the challenge rules (the organisers ship the per-demo replay results with the challenge code, and the human reference is computed from them), and looked at the recordings of the ones that fail or skip stages. Fig.~\ref{fig:demos} shows the three clearest examples; Table~\ref{tab:tasks} has the counts.

\paragraph{LiftTray: the recording stops before the lift.} 54 of the 154 demonstrations fail on replay. They all hold the tray with both hands (3 of 4 stages), but the recording ends with the tray barely raised, about 1~cm against the 2.3~cm the task needs.

\paragraph{StackTwoBlocks: one arm, one cube at a time.} 133 of 141 demonstrations never hold the second cube, so their stage progress is 0.67; only one reaches every stage (Fig.~\ref{fig:demos}). The efficiency reference is the median of these one-arm demonstrations, so a policy that does reach every stage needs a longer path than the reference (1.06~m against 0.80~m in my case).

\paragraph{PackBox: the flaps are pushed shut, never held.} 96 of 99 demonstrations miss stages: they push the flaps over and press them down instead of holding them with the fingers (Fig.~\ref{fig:demos}).

\paragraph{CubeHandover: standing still for longer than the budget.} In two demonstrations nothing moves for the first 20--25~s, so the replay runs out of the 20~s budget before anything happens.

\paragraph{LiftPot: a tilted lift.} 5 of 77 demonstrations fail on replay, and 4 of the successful ones reach only 3 of 4 stages. One demonstration (episode 36) lifts the pot visibly tilted and reaches only 3 of 4 stages.

\paragraph{StackSingleBookShelf: the book is never picked up.} 7 of 73 demonstrations fail on replay. In the one in Fig.~\ref{fig:demos} the recording itself fails: the book stays on the counter and the empty hand goes to the shelf.

\paragraph{PickSingleBookFromTable: the recording ends at the start of the lift.} 9 of 92 demonstrations fail on replay; some of them end with the book held just above the counter.

RotateValve is the only task without problems: all 115 demonstrations succeed with every stage.

\subsection{Suggestions}
\begin{enumerate}[leftmargin=1.4em]
  \item \textbf{Clean demonstrations, and clear requirements.} Most of the problems above are easy to fix: drop or re-record the demonstrations that fail or skip stages, and compute the references only from the ones that reach every stage. Technically, failed or incomplete demonstrations can still be released, just as a separate dataset. While at it, I would also revisit some of the stages themselves. Why does stacking two blocks require holding both of them at some point, and why do the box flaps have to be closed with the fingers? Both look a bit artificial, and a human would not do it that way.
  \item \textbf{Room above the human median.} At the moment a behavior metric gives 1 at the median demonstration and nothing more. Giving points for going below it (faster, smoother, shorter paths than the median human) would keep the strong entries apart; my coordination is 1.000 on every task, for example.
  \item \textbf{Median vs.\ mean.} The policy is scored by its mean over the episodes, but the reference is the human median, and these metrics are skewed. For collisions the median demonstration has 0 on most tasks, while any policy's mean over 50 episodes is a bit above 0 as soon as one episode touches something. So a policy is punished for something the human reference also does now and then. Comparing like with like (mean against mean, or median against median) would fix it.
  \item \textbf{Behavior only from successful episodes.} A failed episode already gets 0 for its outcome. Including its path length and time in the behavior averages punishes it a second time (failures usually wander around until the time runs out); computing the behavior scores only over the successful episodes would be cleaner.
  \item \textbf{Record a bit past completion.} An episode ends at the first successful step, so a tray lifted 2.3~cm for one frame counts. Either ask the goal state to hold for a moment, or simply record the demonstrations 1--2~s beyond completion, so that they show a finished task.
  \item \textbf{Contacts at reset.} In about one PackBox seed in five the robot touches the cabinet during the reset, before the policy does anything. Counting collisions from the first action would keep the safety score about the policy.
  \item \textbf{More episodes per task.} A 50-episode run moves by about $\pm$0.005 RDS at this level. More episodes (cheap in simulation) would make close rankings easier to read, and the final evaluation shows it: the top three entries, within 0.002 of each other on 50 seeds per task, were clearly apart on 1{,}000 (Section~\ref{sec:final}).
  \item \textbf{Harder tasks.} The current tasks are close to saturated, and v2 needs harder ones: more objects and distractors, longer horizons, perturbations and visual randomisation.
  \item \textbf{Recovery as a metric.} People adapt when something goes wrong, and a policy should too. Episodes with perturbations (a nudged object, a dropped grasp) and a score for getting back on track would measure it directly.
  \item \textbf{Synthetic and privileged data.} I think it is a valid approach if you have a good sim-to-real pipeline, but it is less generic otherwise. Either way, I would keep it allowed: scripted or simulator-generated data cannot be told apart from any other data in a submission, so a ban would not be enforceable, and it would also contradict the ``any data and any pretraining'' rule. Harder tasks are a better answer, because they make scripting a less obvious shortcut.
\end{enumerate}

\end{document}